\documentclass[10pt,twocolumn,letterpaper]{article}

\usepackage[accsupp]{axessibility} % Improves PDF readability for those with disabilities.
\usepackage{cvpr}              % To produce the CAMERA-READY version
\usepackage{indentfirst} 
\usepackage{caption}
\usepackage{tabularx}
\usepackage{multirow}
\usepackage{makecell}
\usepackage{wrapfig}
\usepackage{array}  % Required for defining new column types
\usepackage{tabularray}
\newcolumntype{Y}{>{\centering\arraybackslash}X}
\newcolumntype{C}[1]{>{\centering\arraybackslash}m{#1}} % center V and H
\newcolumntype{L}[1]{>{\raggedright\arraybackslash}m{#1}} % align to L and center in V
\newcolumntype{B}[1]{>{\raggedright\arraybackslash}b{#1}}

\usepackage[dvipsnames]{xcolor}

\definecolor{cvprblue}{rgb}{0.21,0.49,0.74}
\usepackage[pagebackref,breaklinks,colorlinks,citecolor=cvprblue]{hyperref}

\def\paperID{*****} % *** Enter the Paper ID here
\def\confName{CVPR}
\def\confYear{2024}

\title{Zero-Shot Monocular Motion Segmentation in the Wild by Combining Deep Learning with Geometric Motion Model Fusion}

\author{
Yuxiang Huang \hspace{1cm} Yuhao Chen \hspace{1cm} John Zelek\\[0.1em]
Vision and Image Processing Lab, University of Waterloo\\[0.1em]
Waterloo, ON, Canada\\
{\tt\small \{yuxiang.huang, yuhao.chen1, jzelek\}@uwaterloo.ca}
}

\begin{document}
\maketitle
\begin{abstract}
Detecting and segmenting moving objects from a moving monocular camera is challenging in the presence of unknown camera motion, diverse object motions and complex scene structures. Most existing methods rely on a single motion cue to perform motion segmentation, which is usually insufficient when facing different complex environments. While a few recent deep learning based methods are able to combine multiple motion cues to achieve improved accuracy, they depend heavily on vast datasets and extensive annotations, making them less adaptable to new scenarios. To address these limitations, we propose a novel monocular dense segmentation method that achieves state-of-the-art motion segmentation results in a zero-shot manner. The proposed method synergestically combines the strengths of deep learning and geometric model fusion methods by performing geometric model fusion on object proposals. Experiments show that our method achieves competitive results on several motion segmentation datasets and even surpasses some state-of-the-art supervised methods on certain benchmarks, while not being trained on any data. We also present an ablation study to show the effectiveness of combining different geometric models together for motion segmentation, highlighting the value of our geometric model fusion strategy.

\end{abstract}    
\section{Introduction}
\label{sec:intro}

Motion segmentation is a fundamental problem in computer vision. It has an essential role in many applications such as action recognition, autonomous navigation, object tracking, and scene understanding in general. The objective of motion segmentation is to divide a video frame into regions segmented by common motions. Motion segmentation becomes particularly challenging when utilizing a single camera that is also moving, as this introduces issues such as degenerate motions, motion parallax, motion on the epipolar plane \cite{Hartley2004}. Existing motion segmentation methods often fails when facing these challenges since they usually rely on only a single motion cue \cite{mohamed_monocular_2021, vertens_smsnet_2017, ramzy_rst-modnet_2019, dave_towards_2019, sekkati_variational_2007, wedel_detection_2009, papazoglou_v_nodate, fragkiadaki_video_2012, keuper_motion_2015}, limiting their effectiveness across the diverse tapestry of real-world environments. While a few deep learning based methods are able to incorporate additional motion cues in an end-to-end manner, their reliance on large annotated datasets and the need for substantial computational resources for training limit their adaptability and application in novel environments \cite{neoral_monocular_nodate, homeyer_moving_nodate}. When facing these challenges, existing methods usually fail to detect the correct motion patterns and also fail to produce coherent segmentation masks for the moving objects. 

In order to overcome these limitations and achieve in-the-wild monocular motion segmentation regardless of motion types and scene structures, it is necessary to have a robust and comprehensive motion model. We draw inspiration from two branches of well studied motion segmentation approaches: and point trajectory based methods and optical flow based methods. These two types of motion cues are not only complementary in nature (long-term vs short-term motion), but they can also be used to derive highly complementary geometric motion models for different motion types and scene structures. Point trajectory based methods, when analyzed using epipolar geometry, will fail if the motion is mainly on the epipolar plane or degenerate (e.g., pure forward motion), but are robust to depth variations, perspective effects and motion parallax. On the other hand, optical flow based methods do not handle these challenges well, but are robust to motions on the epipolar plane. We propose to combine these two motion cues and monocular depth information at the object level using multi-view spectral clustering, to obtain a coherent and comprehensive motion representation of the scene. By doing so, we are able to distinguish a variety of complex object motions (e.g., degenerate motions, motion parallax and non-rigid motion), even in complex scenes. 

In addition to having a comprehensive geometric motion model for effectively analysis of object motions, it is also essential to obtain accurate object proposals for tracking potential moving objects throughout the video. This step is not only fundamental for accurate assessment of object motion, but also vital for generating precise and coherent segmentation masks for moving objects. To accomplish this, we leverage the strong zero-shot ability of the recent computer vision foundation models to identify, segment and track any potential moving objects throughout the video. We then calculate pairwise motion affinity scores for every object pair in the proposal, assessing how well each object-specific motion cue fits its corresponding geometric motion model. The motion affinity scores are used to construct motion affinity matrices, which can be fused by multi-view spectral clustering techniques to obtain the final clustering of objects in different motions

\vspace{-0.1cm}
\begin{figure} [htb] % Use figure* for spanning both columns
    \centering
    \begin{subfigure}{0.48\textwidth}
        \includegraphics[width=\linewidth]{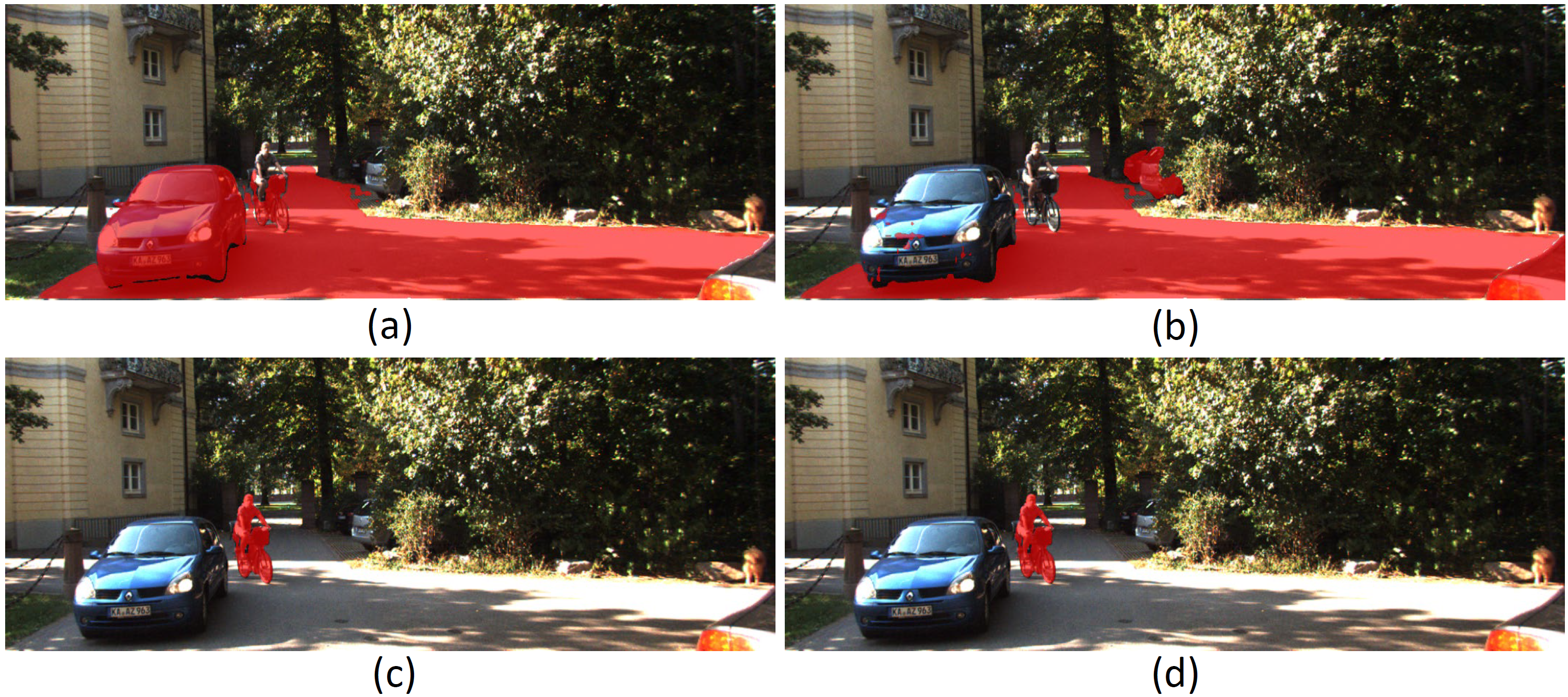}
    \end{subfigure}
    \vspace{-0.5cm}
    \caption{Motion segmentation results from the proposed method using different motion cues on a scene with motion parallax and degeneracy. Motion cues used: (a) point trajectory. (b) optical flow. (c) optical flow + depth. (d) trajectory + optical flow + depth. Using a single motion cue is insufficient to correctly segment out the moving cyclist.}
    \label{fig: Qualitative Segmentation Results}
\end{figure}
\vspace{-0.2cm}

Our method was evaluated on three benchmarks: DAVIS-Moving, YTVOS-Moving \cite{dave_towards_2019}, and an extended version of the KT3DMoSeg dataset we proposed. Our method achieves competitive results on all benchmarks and even surpasses the state-of-the-art supervised method on DAVIS-Moving. 
\section{Related Work}

Research on monocular motion segmentation has been ongoing for several decades, leading to varying interpretations of the problem among different studies. Commonly, it is defined as the process of dividing a video frame into regions that share similar motions. Alternatively, many studies also approach motion segmentation as the task of clustering predefined feature point trajectories across two or more video frames based on their distinct motions. In this paper, we focus on performing motion segmentation directly from input video frames, with the goal of segmenting entire moving objects, including those exhibiting multiple rigid motions. This approach aims at attaining a high-level understanding of the scene. 

Monocular motion segmentation can be broadly divided into three distinct categories, each defined by the type of motion cues utilized. The first group consists of optical flow based methods, which rely on optical flow as their primary source of motion information \cite{sekkati_variational_2007, cremers_detection_2009, papazoglou_v_nodate, leibe_its_2016, vertens_smsnet_2017, bideau_best_2018, siam_modnet_2018, ramzy_rst-modnet_2019, dave_towards_2019, leal-taixe_moa-net_2019, mohamed_monocular_2021, meunier_em-driven_2023}. The second group includes the feature point trajectory based methods, which rely exclusively on motion information derived from manually corrected feature point trajectories throughout the video \cite{delong_fast_2010, isack_energy-based_2012, hutchison_object_2010, brox_object_2010, elhamifar_sparse_2013, ochs_segmentation_2014, lai_motion_2017, xu_motion_2018, vedaldi_usage_2020}. The last category comprises fusion-based methods, which combine multiple types of motion cues as well as appearance cues, to enhance the segmentation results \cite{neoral_monocular_nodate, homeyer_moving_nodate, huang_dense_2024}.

% These techniques predominantly employ deep learning frameworks to train neural networks for the end-to-end fusion of diverse motion cues. To date, exploration within this category has been limited to deep learning methods, with non-deep learning strategies yet to be investigated.

% It is also important to clarify the distinction between motion segmentation and video object segmentation (VOS). While VOS aims to isolate moving objects primarily in the foreground, motion segmentation seeks to identify and segment any object exhibiting independent motion, irrespective of its position in the foreground or background. This distinction underscores the more general scope of motion segmentation in analyzing video content.

\subsection{Optical Flow Based Methods}
Optical flow based methods can be further categorized into traditional and deep learning based methods. Traditional methods \cite{sekkati_variational_2007, cremers_detection_2009, papazoglou_v_nodate, leibe_its_2016} rely on optical flow masks input, and produce a pixel-wise segmentation mask indicating different motion groups. These methods usually adopt iterative optimization approaches or statistical inference techniques to estimate the motion models and motion regions simultaneously. In contrast, numerous deep learning based methods \cite{ramzy_rst-modnet_2019, faisal_epo-net_2020, bosch_deep_2021, cao_learning_2019, mohamed_monocular_2021} use a CNN encoder to extract motion cues from optical flow and uses a decoder to produce the final segmentation. More advanced deep learning models use two CNN encoders -- one to extract motion information from optical flow and the other one to extract appearance features directly from the video frame -- to enhance the segmentation performance. However, deep learning methods often require a large amount of training data and do not generalize well to novel scenes. 

In general, optical flow based methods perform well on scenes without strong depth variations or motion parallax. However, if the scene contains these elements (e.g. road scenes), these methods will fail to distinguish if a part of the image is moving independently or is just at a different depth from its surroundings, because the motion flow vectors projected to a 2D image from the 3D space are determined by both the depth and the screw motion of the object \cite{mitiche_computer_2014}. Additionally, strong brightness changes in the video will also adversely affect the performance of optical flow based methods since optical flow calculation is based on the brightness constancy constraint, which will be violated under strong brightness change.

\subsection{Point Trajectory Based Methods}

In contrast to the other two categories, point trajectory based methods produce clusters of key points that represent various motion patterns, rather than providing full dense segmentations. These techniques can be further divided into two-frame and multi-frame methods. Two frame methods \cite{delong_fast_2010, isack_energy-based_2012, barath_progressive-x_2019} usually determine motion parameters by solving an iterative energy minimization problem of finding a certain number of geometric models (e.g., fundamental matrices) on a set of matched feature points, to minimize an energy function that evaluates the quality of the overall clustering of correspondences. Multi-frame based methods, on the other hand, usually analyze manually adjusted trajectory points from a dense optical flow tracker and often employ spectral clustering on affinity matrices. These matrices are generated through geometric model fitting \cite{lai_motion_2017, xu_motion_2018, vedaldi_usage_2020, huang_motion_2023, huang_unified_2023}, subspace fitting \cite{elhamifar_sparse_2013, rao_motion_2010, tron_benchmark_2007, vidal_subspace_2011}, or pairwise motion affinities derived from spatio-temporal motion cues and appearance cues \cite{brox_object_2010, ochs_segmentation_2014}. 

The efficacy of point trajectory based methods is heavily influenced by the chosen motion model and the precision of point correspondences. There is not a single motion model that can capture motion similarities across all types of motions. In search of a better motion model, \cite{vedaldi_usage_2020} uses trifocal tensor to analyze point trajectories. Trifocal tensor is more robust to noises and is able to distinguish motions on the epipolar plane, but it is harder to optimize and prone to failure when the three cameras are close to being colinear \cite{Hartley2004}, which can often happen on road scenes. \cite{xu_motion_2018, jiang_what_2021} proposed geometric model fusion techniques to combine different geometric models, but they still fail to produce coherent and consistent segmentations on complex scenes. Moreover, most existing methods depend on manually refined point correspondences and struggle to effectively manage outliers.

\begin{figure*}[t]  % Use figure* for spanning both columns
    \centering
    \includegraphics[width=\linewidth]{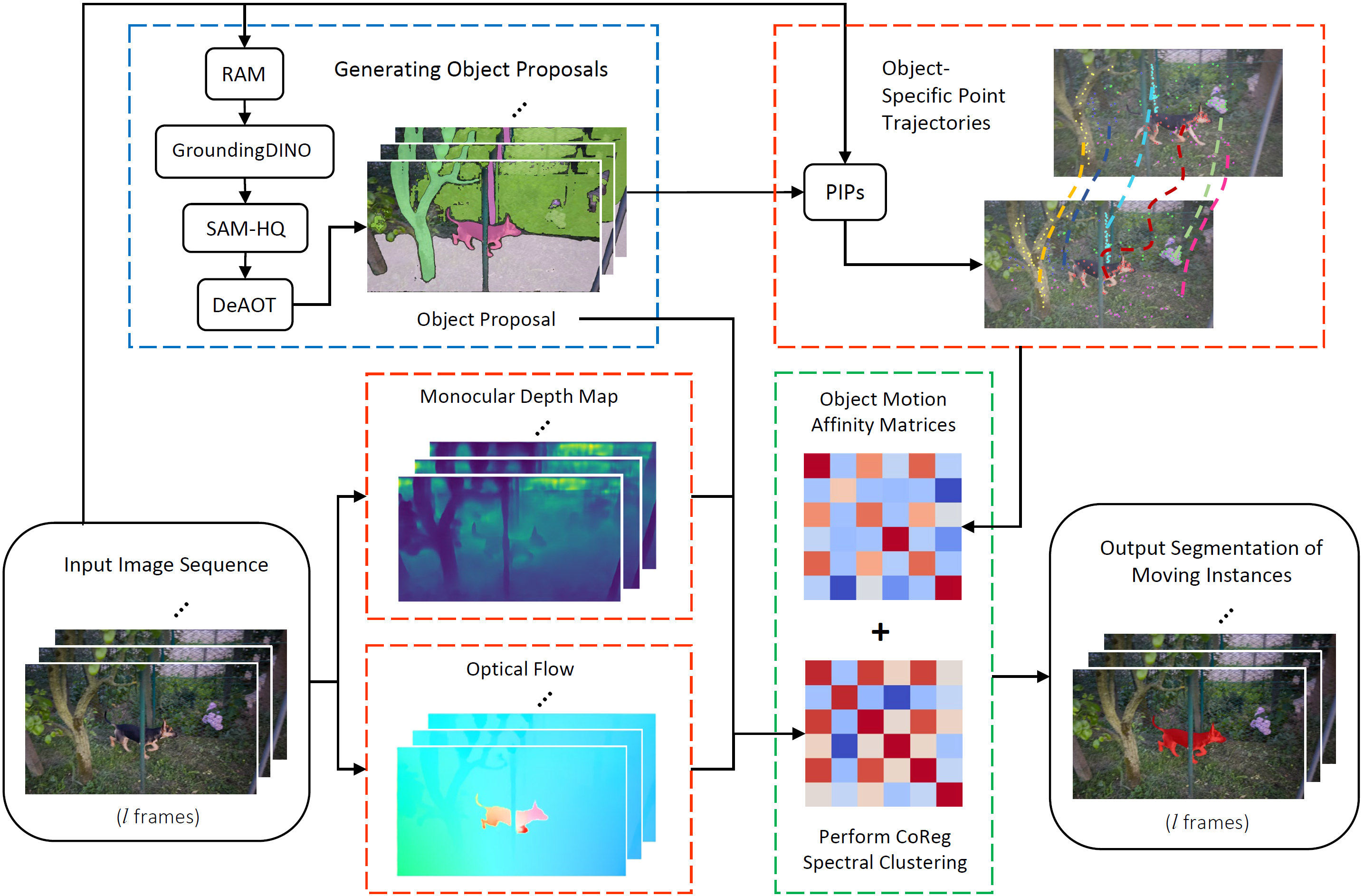} 
    % \caption{(a) The instance segmentation mask obtained from the preprocessing pipeline on the Seq095_Clip01 sequence from the KT3DMoSeg dataset. (b) The initial grouping of trajectories based on the instance segmentation -- trajectories on the same object or background are assigned the same initial label}
    % \vspace{-0.4cm}
    \caption{Our Motion Segmentation Pipeline. Our method can be summarized to three main steps: 1) given a sequence of video frames, we produce an object proposal by automatically detecting, segmenting and tracking common objects in the video. 2) we compute object-specific point trajectories, optical flow and monocular depth maps for every frame. 3) we compute pairwise object motion similarity scores using two motion models (one based on point trajectories and the other based on optical flow and depth map), and use them to construct two motion affinity matrices. The two matrices are fused using multi-view spectral clustering to cluster objects into different motion groups.}
    \label{fig: Motion Segmentation Pipeline}
    \vspace{-0.1cm}
\end{figure*}

\subsection{Fusion Based Methods} 
Recent research in motion segmentation have introduced several innovative approaches that leverage a combination of motion cues for improved motion segmentation accuracy. Notably, \cite{neoral_monocular_nodate} integrates optical flow masks and monocular depth maps through a fusion module in their neural network, facilitating end-to-end training. By adopting a semi-supervised training strategy, this approach has set new benchmarks in monocular dense motion segmentation across various datasets. Another study \cite{homeyer_moving_nodate} explored the impact of utilizing different combinations of motion cues, such as optical flow, depth map, and scene flow, on motion segmentation performance, achieving state-of-the-art results on the KITTI and DAVIS datasets. However, this approach is fully supervised, which requires an extensive amount of training data and computational power. Furthermore, its ability to generalize across a wider range of benchmarks remains unverified. In our earlier work \cite{huang_dense_2024}, we introduced an interpretable geometric model that merges optical flow with monocular depth maps for zero-shot motion segmentation. Despite these efforts, a noticeable performance gap persists between our method and the state-of-the-art supervised or semi-supervised techniques, showing the insufficiency of relying solely on a single geometric model to achieve optimal results in motion segmentation. 

Despite these recent research, no existing method has yet to combine the two complementary and most commonly explored motion cues in motion segmentation: point trajectory and optical flow. This paper seeks to fill this gap, demonstrating how combining these motion cues with well-crafted geometric motion models can lead to state-of-the-art zero-shot monocular motion segmentation.

\section{Methodology}

We propose a zero-shot monocular motion segmentation approach that uses both object appearance information and a combination of epipolar geometry and optical flow based geometric motion models to perform in-the-wild motion segmentation without any assumptions of the motion or the scene that may appear in the video.

Our segmentation pipeline begins by identifying initial segmentation of the background and common objects within the scene through foundational models, followed by continuous tracking of these objects across the video sequence. For every object in each frame, we gather a collection of object-specific trajectory points, an optical flow mask, and a monocular depth map. Subsequently, we construct two distinct geometric motion models for each scene object: one via fundamental matrix fitting using point trajectories and the other via fitting optical flow and a depth map to our proposed parametric equations. By fitting each object's motion models on every other object and analysing the residuals of the model fitting, we are able to derive two pairwise affinity scores between every pair of objects, from which we can construct two motion affinity matrices for the two types of motion models respectively. Lastly, we fuse the two affinity matrices using co-regularized multi-view spectral clustering to obtain the final segmentation. Figure \ref{fig: Motion Segmentation Pipeline} shows a diagram of the motion segmentation pipeline.

\subsection{Generating Object Proposals}
In order to identify all motions in a video sequence at object level, we use the same method as proposed in \cite{huang_dense_2024} to identify, segment and track each prominent object across the video. This is accomplished by integrating foundation models for object recognition (RAM) \cite{zhang_recognize_2023}, detection (Grounding DINO model) \cite{liu_grounding_2023}, segmentation (SAM-HQ) \cite{ke_segment_2023}, and tracking (DeAOT) \cite{yang_decoupling_2022}. This video preprocessing pipeline for automatic object proposal generation is inspired by and improved upon Segment and Track Anything (SAMTrack) \cite{cheng_segment_2023}. Comparing to SAMTrack, our video preprocessing module combines these foundation models to segment and track objects automatically, bypassing the need for manual text prompts by initiating our pipeline with the Recognize Anything Model to automatically detect common objects in the initial video frame. Our object proposal generation pipeline involves: 1) Automatically identifying common objects in the video's first frame using the Recognize Anything Model; 2) Generating object bounding boxes with the Grounding DINO model; 3) Producing instance segmentation masks for the initial frame via the SAM-HQ model, applying non-max suppression to refine the results; 4) Employing the DeAOT tracker to track each object's segmentation mask throughout the video. To accommodate new objects appearing mid-sequence, we segment the video into sections of $l$ frames, repeating the above process for each section. The choice of $l$ varies depending on the video's dynamics and the frequency of new objects entering the scene. Videos with higher dynamics and frequent entry of new objects mid-sequence are better suited to a reduced segment length $l$.

\subsection{Object-Specific Motion Cues}
Once we have an object proposal for every frame of the video, we will then obtain object-specific motion cues for every object in the object proposal. We propose to use point trajectories, optical flow and monocular depth map automatically generated by off-the-shelf networks as motion cues, in order to model objects' motions in two complementary ways. 
\subsubsection{Object-Specific Point Trajectories}
A set of sparse point trajectories is generated for every object using PIPs \cite{avidan_particle_2022}. PIPs is a state-of-the-art point tracker which tracks individual pixels given their initial locations in a video frame. A mixture of Shi-Tomasi \cite{jianbo_shi_good_1994} and K-Medoids \cite{park_simple_2009} sampling method is used to obtain the initial pixels from each object as it showed good experimental results from previous works in similar tasks \cite{rajic_segment_2023}. These tracked pixels can be used as object-specific feature points to fit fundamental matrices for every object in frame pairs to describe their motions. One limitation of PIPs is that does not handle occlusion well if the tracked video is more than 8 frames. To overcome this issue, we check for every point if it is inside its corresponding object's mask area every 8 frames. If not, we remove that point and sample a new point inside the object's mask. We also remove any point that is near the edge of the frame since the tracking accuracy of PIPs drops significantly in this case. 

\begin{figure*} [t!]  % Use figure* for spanning both columns
    \centering
    \includegraphics[width=\textwidth]{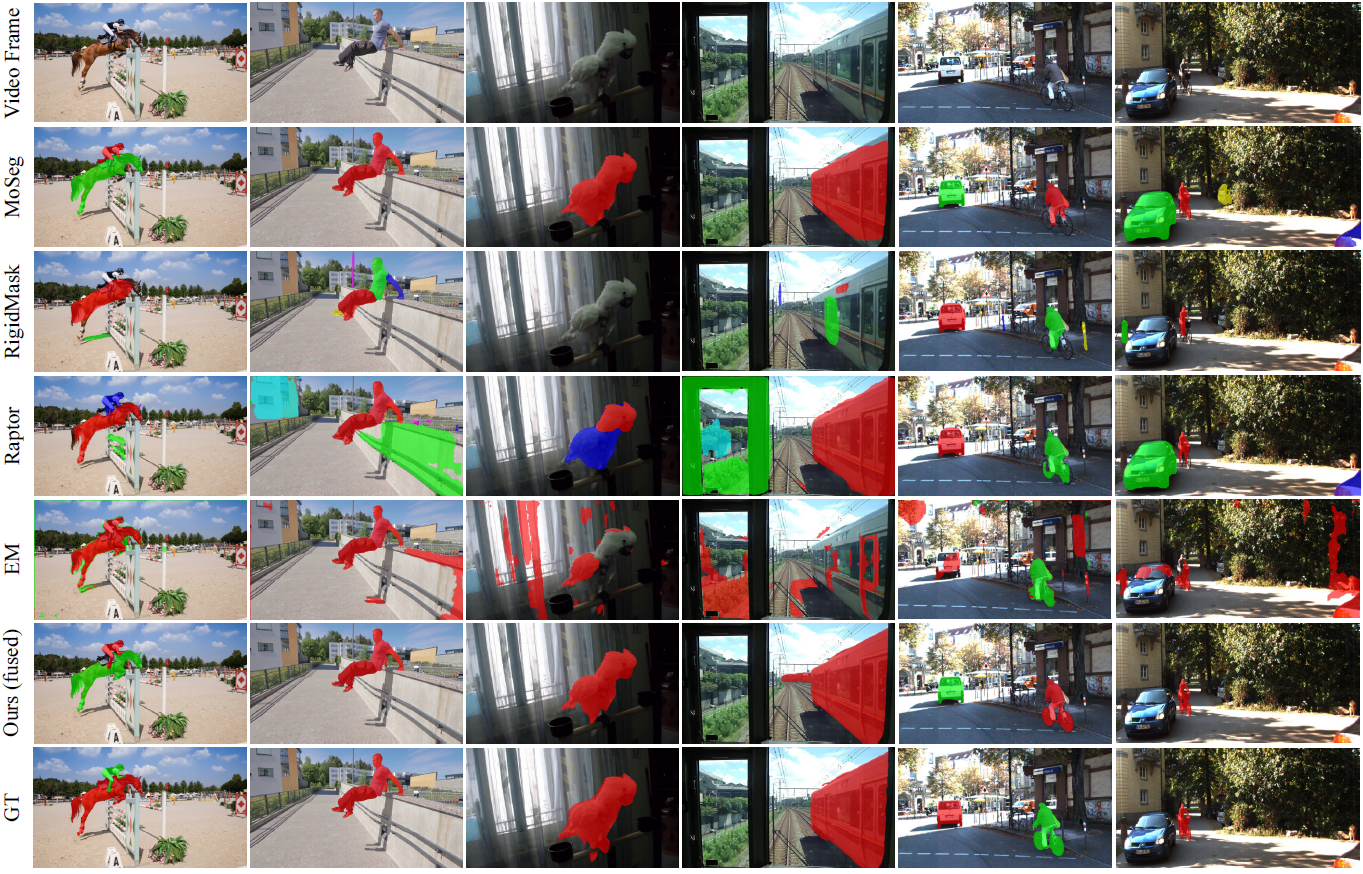} 
    \vspace{-0.6cm}
    \caption{Qualitative results of different methods on DAVIS-Moving (row 1, 2), YTVOS-Moving (row 3, 4) and the extended KT3DMoSeg (row 5, 6) datasets. MoSeg often mistakenly label static objects as dynamic when there is degenerate camera motion. RigidMask fails to detect or coherently segment objects with non-rigid motions. Similarly, Raptor also has these problems, although to a lesser extent overall. Our method, despite being zero-shot, performs well when facing these challenges.}
    \label{fig: MoSeg Results}
    \vspace{-0.1cm}
\end{figure*}

\subsubsection{Object-Specific Optical Flow and Depth Map}
We also generate a dense optical flow mask and a monocular depth map for every frame, from which we can extract object-specific optical flow vectors and depth maps. We use a state-of-the-art optical flow estimator \cite{sun_disentangling_2022} to obtain optical flow, and a state-of-the-art monocular depth estimator, DINOv2 \cite{oquab_dinov2_2023}, to extract the depth maps. We use monocular depth estimation to estimate the scene depth from a single frame since our goal is to perform motion segmentation from a moving monocular camera. DINOv2 outputs a relative depth map, which is sufficient for our application. Our experiment shows improved results when both optical flow and depth map are used to compute the motion model, comparing to only optical flow. We show how a depth map can be used to improve the motion model based solely on optical flow in the next section.

\vspace{-0.2cm}
\begin{table*}[h!t!b!p!]
  \renewcommand{\arraystretch}{1.5} % spacing between rows
  \centering
  \begin{tabularx}{\textwidth}{|C{0.6cm} |L{2.5cm} ||L{2cm} |YYY |YYY |YYY|}
    \hline
    \multirow{2}{*}{\raggedright \makecell{\textbf{Exp.}}} &
    \multirow{2}{*}{\raggedright \makecell{\textbf{Method}}} & 
    \multirow{2}{*}{\centering \makecell{\textbf{Training}}} & 
    \multicolumn{3}{c|}{\textbf{DAVIS-Moving}} & 
    \multicolumn{3}{c|}{\textbf{YTVOS-Moving}} & 
    \multicolumn{3}{c|}{\textbf{KT3DInsMoSeg}} \\
    % \textbf{Exp.} &  \textbf{Method} & \textbf{Training} &
    \textbf{} &  \textbf{} & \textbf{} &
    \textbf{Pu} & \textbf{Ru} & \textbf{Fu} & 
    \textbf{Pu} & \textbf{Ru} & \textbf{Fu} & 
    \textbf{Pu} & \textbf{Ru} & \textbf{Fu} \\
    \hline
    a & MoSeg \cite{dave_towards_2019} & Supervised & \textbf{78.30} & 78.80 & \underline{78.10} & \textbf{74.50} & \underline{66.40} & \textbf{66.38} & 63.73 & 78.24 & 66.85 \\
    \hline
    a & Raptor \cite{neoral_monocular_nodate} & \multirow{2}{*}{\raggedright \makecell{Supervised \\ Features}} & 75.90 & 79.67 & 75.93 & \underline{64.43} & 60.94 & 60.35 & 71.52 & \textbf{88.27} & \textbf{75.82} \\
    a & RigidMask \cite{yang_learning_2021} & \textbf{} & 59.03 & 49.89 & 50.01 & 29.88 & 17.88 & 18.70 & 65.14 & 83.34 & 70.91 \\
    \hline
    a & EM \cite{meunier_em-driven_2023} & Unsupervised & 58.42 & \underline{83.48} & 64.24 & 44.52 & 40.33 & 37.12 & 42.85 & 58.71 & 44.03 \\
    \hline
    \hline
    a+b & \textbf{Ours (fused)} & \multirow{5}{*}{\raggedright \makecell{Zero-Shot \\ (no training)}} & \underline{78.27} & 81.58 & \textbf{79.40} & 64.12 & 61.10 & \underline{60.62} & \textbf{72.93} & 71.02 & \underline{71.89} \\
    b & Ours (OC+depth) & \textbf{} & 71.53 & 75.66 & 73.18 & 63.54 & 58.94 & 56.06 & 48.04 & 61.54 & 49.26 \\
    b & Ours (OC) & \textbf{} & 58.25 & 59.22 & 57.08 & 61.79 & 54.64 & 53.74 & 36.44 & 39.97 & 34.78 \\
    b & Ours (trajs) & \textbf{} & 65.99 & 75.51 & 68.47 & 54.67 & 52.92 & 50.05 & 42.31 & 73.66 & 45.24 \\
    b & Ours (base) & \textbf{} & 43.17 & \textbf{86.24} & 52.12 & 48.49 & \textbf{73.01} & 50.82 & 38.97 & 70.97 & 43.37 \\
    \hline
  % \end{tabular}
   \end{tabularx}
  \caption{Performance of our method and state-of-the-art motion segmentation methods (\textbf{Exp.} a) on the DAVIS-Moving, YTVOS-Moving validation datasets and the KT3DInsMoSeg dataset, as well as ablation study results (\textbf{Exp.} b). The best result for each metric is in bold and the second best result is in underscore. Our method overall performs the best on DAVIS-Moving and second best on both YTVOS and KT3DInsMoSeg, despite not being trained on any data. Our method also significantly surpasses the state-of-the-art unsupervised motion segmentation method \cite{meunier_em-driven_2023}.}
  \label{table: Quantitative Results and Ablation Study}
  \vspace{-0.2cm}
\end{table*}

\subsection{Geometric Motion Model Fitting}
After obtaining object-specific point trajectories, optical flow vectors and depth maps, for each frame pair, we compute two geometric motion models of each object based on epipolar geometry and optical flow respectively, to model its motion throughout the video. To compute the epipolar geometry based motion models using point trajectories, we compute a fundamental matrix of each object between every {\it f} frames by solving ${p'T F p = 0}$ using the eight-point algorithm with RANSAC \cite{fischler_random_1981}, where $p$ and $p'$ are the normalized 2D homogeneous coordinates of the same tracked point in the two frames. If a degenerate case is encountered for the fundamental matrix, we do not use it. 

For the optical flow and depth based motion model, we use the same motion model proposed in our earlier work \cite{huang_dense_2024}. We refine the Longuet-Higgins and Pruzdny model equation \cite{longuet-higgins_interpretation_1980} to address rigid object motion, adapting it to include depth information without needing exact pixel depth, a common limitation in practice. Instead of relying on the original model, which is impractical due to unknown absolute pixel depth, we propose a linearized version incorporating relative depth from DINOv2, making it more applicable to real-world scenarios with varying depths. This approach, while using both optical flow and depth data, simplifies the motion model to the following linear equations:

\vspace{-0.2cm}
\begin{equation} \label{eq:2}
\begin{split}
u = a + b \frac{1}{z} - c \frac{x}{z} -dy + ex^2 - fxy
\\[1ex]
v = g + h \frac{1}{z} - c \frac{y}{z} -dx + exy + fy^2
\end{split}
\end{equation}
\vspace{-0.2cm}

This motion model aims to cluster different motions rather than calculate exact screw motions, sidestepping scale uncertainties and making it theoretically sound without requiring specific camera intrinsics. For consistency reasons, we still refer to this motion model as the  ''optical flow motion model", although it uses both optical flow vectors and pixel depth maps. 

% full quadratic motion model with 12 parameters to model the instantaneous object screw motion:
% \begin{equation} \label{eq:1}
% \begin{split}
% f(x, y) = & (a + b x + c y + d x^2 + e xy + f y^2, \\
% & g + h x + i y + j x^2 + k xy + l y^2)
% \end{split}
% \end{equation}

% where {\it (x, y)} is the 2D coordinates of the pixels relative to the image center. Since we already have the instance optical flow field, we can obtain the following equation:
% \begin{equation} \label{eq:2}
% \begin{split}
% f(x, y) = (u, v)
% \end{split}
% \end{equation}

% where $(u, v)$ is the optical flow vector of the pixel. We fit the function \ref{eq:1} above to the the optical flow vectors of every object and solve for the 12 parameters representing the object motion model by optimizing the mean squared error. We use this specific motion model as it's a simplified version of the classic Longuet-Higgins and Pruzdny model equations \cite{longuet-higgins_interpretation_1980}, which model's the instantaneous screw motion of rigid objects at arbitrary depth. Since it's not possible to solve for the depth of each pixel, this motion model assumes the objects' depths are only slightly different. It was shown to perform well on scenes with limited motion parallax \cite{meunier_em-driven_2023}, nevertheless, it often fails when there is strong motion parallax and depth variations. 

\subsection{Constructing Motion Affinity Matrices}
After all fundamental matrices and optical flow motion models are computed, each object will have a fundamental matrix between every $p$ frames and an optical flow motion model between every two frames. By fitting every object's trajectory points,  optical flow vectors and depth maps to every other object's fundamental matrix and optical flow motion model on the same frame pair, we can obtain the residuals of every object to all other objects' motion models respectively. We use Sampson distance \cite{Hartley2004} as the residual for the fundamental matrix and mean squared error for the optical flow motion model. Assuming there are {\it k} objects in the scene, for the {\it i}-th object  at the {\it m}-th frame pair, we obtain the following residual vectors under the fundamental matrix and optical flow motion models:
\begin{equation*} \label{eq:3}
\setlength{\jot}{10pt} % line spacing in the environment
\begin{split}
{\pmb r_{o}}_i^m = [{r_{o}}_{i,1}^m, {r_{o}}_{i,2}^m, ..., {r_{o}}_{i,k}^m], 
\\
{\pmb r_{f}}_i^m = [{r_{f}}_{i,1}^m, {r_{f}}_{i,2}^m, ..., {r_{f}}_{i,k}^m]
\end{split}
\end{equation*}
where ${r_{o}}_{i,k}^m$ is the mean residual for fitting the parametric motion model of object $i$ on the optical flow vectors of object $k$ between frames $m$ and $m + 1$, and ${r_{f}}_{i,k}^m$ is the mean Sampson error for fitting the fundamental matrix of object $i$ on the trajectory points of object $k$ between frames $m$ and $m + p$. We construct two affinity matrices encapsulating the pairwise motion affinities between each pair of objects using a modified version of ordered residual kernal (ORK) \cite{chin_ordered_2009}. Specifically, for each object, we sort its residual vectors in ascending order and define a threshold to select the smallest {\it t}-th residual as inliers. We define ${\pmb c_{i}} = \{0,  max(t-n_i, 0)\}^K$ as an inlier score vector, where $n_i$ is the rank of object $k$ in the residual vector of object $i$, penalizing different inlier distributions between objects. The pairwise motion affinity between objects $i$ and $j$ can thus be computed as ${\pmb a_{ij} = \pmb c_{i}^\intercal \pmb c_{j}}$, which denotes a weighted co-occurrence score between two objects as inliers of all motion models. Our proposed weighted ORK is robust to outliers and makes the affinity matrix more adaptive to different scenes by reducing the need to set scene specific inlier thresholds. 
% We construct two affinity matrices encapsulating the pairwise motion affinities between each pair of objects using a modified version of ordered residual kernal (ORK) \cite{chin_ordered_2009}. More specifically, for each object, we sort its residual vectors in ascending order and define a threshold to select the smallest {\it t}-th residual as inliers. We define ${\pmb c_{i}} = \{0, 1\}^K$ as the inlier mask to denote if an object i is an inlier for each of the {\it K} motion models, and the pairwise motion affinity between objects {\it i} and {\it j} can be computed as ${\pmb a_{ij} = \pmb c_{i}^\intercal \pmb c_{j}}$, which denotes the co-occurrence between two objects as an inlier of all motion models. ORK is robust to outliers and makes the affinity matrix more adaptive to different scenes by reducing the need to set scene specific inlier thresholds. 

\subsection{Co-Regularized Multi-view Spectral Clustering}
After constructing the affinity matrices, we normalize them using row normalization \cite{von_luxburg_tutorial_2007} and adapt co-regularized multi-view spectral clustering \cite{kumar_co-regularized_2011} to fuse the two affinity matrices together. With the number of motion groups in the scene given as an input, we are able to obtain the final clustering of moving objects. Co-regularized multi-view spectral clustering uses an regularization term to encourage consensus between different views and is shown to perform well on fusing multiple geometric models for a consistent representation of motion information \cite{xu_motion_2018}.

\section{Experiments}

Our method is tested on three benchmarks: DAVIS-Moving, YTVOS-Moving and the extended KT3DMoSeg. We first briefly introduce these datasets, then show both quantitative and qualitative comparisons between our method and other state-of-the-art methods. Lastly, we present an ablation study to compare the performance of each individual motion models and the fused motion model.

\subsection{Datasets}
DAVIS-Moving and YTVOS-Moving are both proposed by \cite{dave_towards_2019} as datasets for generic instance motion detection and segmentation. DAVIS-Moving and YTVOS-Moving are subsets of the DAVIS 17 dataset \cite{pont-tuset_2017_2018} and the YTVOS dataset \cite{ferrari_youtube-vos_2018}, where all moving instances in the video sequence are labeled and no static objects are labeled. These two recently proposed datasets are very challenging due to their diverse object classes, occlusions and non-rigid motions.

In addition to these two datasets, we also evaluate our method on an extended version of the KT3DMoSeg dataset. The original KT3DMoSeg dataset \cite{xu_motion_2018} is designed to test point trajectory based motion segmentation methods on complex road scenes. It contains manually corrected point trajectories on selected moving instances in road scenes and includes significant degenerate motions and depth variation. In order to test the performance of our method in such environments, we extend the KT3DMoSeg dataset by adding a pixel-level segmentation mask to every moving instance in the scene. We refer to this extended dataset as the KT3DInsMoSeg dataset in the following sections.

\subsection{Results and Discussion}

Our method's performance is evaluated using \textit{precision} (Pu), \textit{recall} (Ru), and \textit{F-measure} (Fu) proposed in \cite{dave_towards_2019} which penalizes false positives. The \textit{F-measure} combines both \textit{precision} and \textit{recall} and indicates the method's overall performance. Table \ref{table: Quantitative Results and Ablation Study} shows quantitative results of our method and other state-of-the-art methods on the three benchmarks. Despite no training, our approach excels on the DAVIS-Moving dataset, outperforming fully-supervised methods, ranks second on the YTVOS-Moving dataset, closely surpassing Raptor \cite{neoral_monocular_nodate}, and secures a similar position on the KT3DInsMoSeg dataset, trailing only behind Raptor. Our method also significantly surpasses EM \cite{meunier_em-driven_2023}, which is the state-of-the-art unsupervised multi-label motion segmentation method.  

We also qualitatively compare our method with these methods and show the results in \ref{fig: MoSeg Results}. Results indicate our method's superiority in identifying static and moving objects across various scenes, notably in complex scenarios where other methods fail, such as in scenes with degenerate motions or complex object contours. Our technique demonstrates robust performance across all datasets, showing its effectiveness in accurately grouping motions and outperforming existing methods in challenging conditions.

One primary limitation of the proposed method is its inference speed. Despite being a zero-shot approach that requires no training, the method integrates multiple computer vision foundation models, as well as the neural networks for feature point tracking and optical flow estimation. Consequently, such integration significantly slows the method's processing speed, making it only suitable to be applied on pre-recorded videos. Another limitation is the requirement for a known ground truth number of motions in the scene to achieve optimal results, inherent to the use of spectral clustering. Although this issue can be mitigated by employing various model selection methods \cite{huang_unified_2023, von_luxburg_tutorial_2007}, such adjustments typically result in a slight degradation of performance.

\vspace{-0.2cm}
\begin{figure*} [t!]  % Use figure* for spanning both columns
    \centering
    \includegraphics[width=\textwidth]{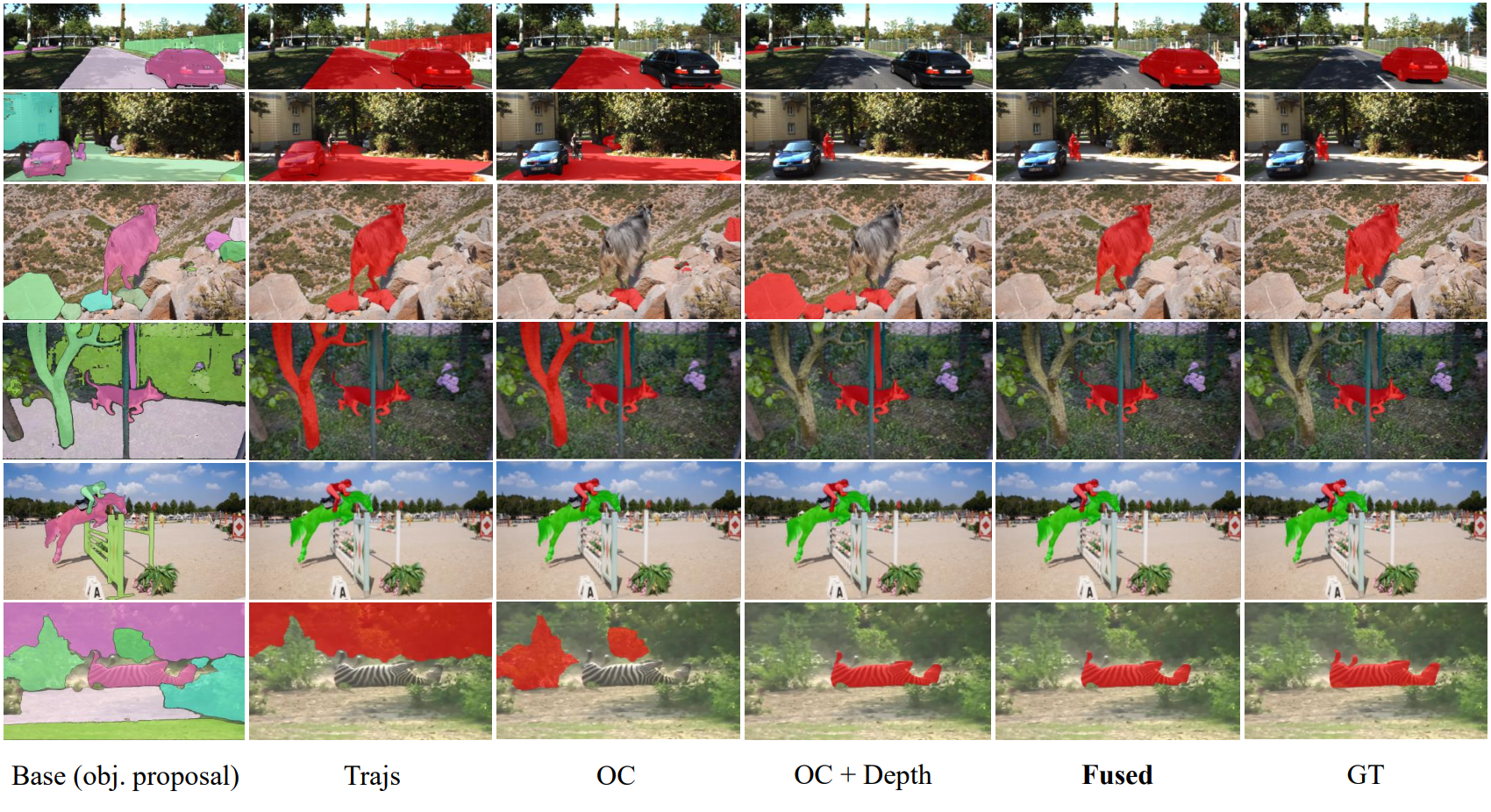} 
    % \caption{(a) The instance segmentation mask obtained from the preprocessing pipeline on the Seq095_Clip01 sequence from the KT3DMoSeg dataset. (b) The initial grouping of trajectories based on the instance segmentation -- trajectories on the same object or background are assigned the same initial label}
    \vspace{-0.8cm}
    \caption{Qualitative comparison of different motion models on different scenes. Pure optical flow based motion model (OC) suffers on scenes with objects at varying depths. Combining optical flow with depth information (OC + Depth) only alleviates this problem to some extent. Pure point trajectory based motion model (Trajs) suffers from motions near the epipolar plane and inaccurate trajectory estimation. Motion model fusion solves these problem by combining the advantages of both motion models and outperforms any single model.}
    \label{fig: Ablation Study}
    \vspace{-0.3cm}
\end{figure*}

\subsection{Ablation Study}

We present both quantitative (Table \ref{table: Quantitative Results and Ablation Study}, Exp. b) and qualitative (Figure. \ref{fig: Ablation Study}) comparisons between different individual motions models and the fused motion model for their performances on the three benchmarks.

We found that on both DAVIS-Moving and KT3DInsMoSeg datasets, our model fusion technique (fused) is able to significantly boost the Fu score comparing to using only a single model, while on YTVOS-Moving, the Fu score only had a relatively small increase. Upon further inspection, we discovered this could be attributed to some motion labels in the YTVOS-Moving dataset actually being mostly static throughout the video sequence. Since our method clusters moving objects purely using motion cues, it groups these objects together with the background as expected. Additionally, the YTVOS-Moving dataset also contains videos with significant camera zooming, which violates the geometric assumptions of both our motion models. Our motion model fusion technique is able to achieve better results than any single motion model on all three datasets, showing its effectiveness.

We also show the motion segmentation performance of our pipeline under conditions where only partial motion cues are used. Specifically, we present results obtained from two different types motion affinity matrices, which are computed using two different motion models: one solely based on the optical flow motion model (OC), and another that combines optical flow with monocular depth information (OC + depth). The optical flow based motion model is obtained from \cite{meunier_em-driven_2023}, which is a state-of-the-art unsupervised method using only optical flow as input. The motion model combining optical flow and depth is proposed by \cite{huang_dense_2024}, which is a direct improvement on \cite{meunier_em-driven_2023}. Results show that the motion model based on a combination of optical flow and depth (OC + depth) outperforms OC by a large margin in all three metrics on both DAVIS-Moving and KT3DInsMoSeg, while having similar results on YTVOS-Moving. 

Both Point trajectory based (trajs) and optical flow based motion models perform poorly on the KT3DInsMoSeg dataset, potentially due to significant motion degeneracy (e.g., forward motion) and depth variations on road scenes. Incorporating depth information in this case proves to be an effective way to reduce motion ambiguity for the optical flow based motion model, boosting its F-score from 34.78\% to 49.26\%. Fusing the combined (OC + depth) motion model with the epipolar geometry based point trajectory motion model significantly enhances the performance in this case.

\section{Conclusion and Future Work}
We propose the first zero-shot monocular motion segmentation approach that achieves state-of-the-art performance. Our method combines the advantages of both deep learning and multiple geometric approaches, resulting in a zero-shot motion segmentation approach that performs geometric motion model fusion on object proposals. We compare the performances of the fused motion model and each individual motion model, and observe a significant performance improvement for the fused motion model, showing the effectiveness of the proposed geometric motion model fusion technique. Even though our method is zero-shot, experiments show that our method is better than many state-of-the-art methods and highly competitive with others. 

Future research could pursue two promising directions: First, the integration of additional motion models, such as the trifocal tensor \cite{Hartley2004}, may further improve the motion segmentation performance. Second, developing methods to effectively incorporate both types of motion affinity measures into the loss function could enable end-to-end, self-supervised training of a motion segmentation network, potentially achieving substantial improvement in inference speed. 
\clearpage

\clearpage
{
    \small
    \bibliographystyle{ieeenat_fullname}
    \bibliography{main}
}


\end{document}